\documentclass[letterpaper, 10 pt, conference]{ieeeconf}  

\IEEEoverridecommandlockouts                              

\makeatletter
\newif\if@restonecol
\makeatother

\usepackage{cite}
\usepackage{amsmath,amssymb,amsfonts}
\usepackage{algorithmic}
\usepackage{graphicx}
\usepackage{textcomp}
\usepackage{xcolor}
\usepackage{booktabs}
\usepackage{bbding}
\usepackage{tikz}
\usepackage{bm}
\usepackage[ruled,linesnumbered,noline]{algorithm2e}

\title{\LARGE \bf
MODUR: A Modular Dual-reconfigurable Robot
}

\author{Jie Gu$^{1}$, Tin Lun Lam$^{2,3,\dag}$, Chunxu Tian$^{1}$, Zhihao Xia$^{1}$, Yongheng Xing$^{1}$, Dan Zhang$^{1,4,\dag}$ 
\thanks{*This work was supported by the National Nature Science Foundation of China (grants 52305012).}
\thanks{$^{1}$Institute of AI and Robotics, Academy for Engineering \& Technology, Fudan University, Shanghai 200433, China.}
\thanks{$^{2}$School of Science and Engineering, The Chinese University of Hong Kong, Shenzhen, China.}
\thanks{$^{3}$Shenzhen Institute of Artificial Intelligence and Robotics for Society.}
\thanks{$^{4}$Department of Mechanical Engineering, The Hong Kong Polytechnic University, Hung Hom.}%
\thanks{$^{\dag}$Corresponding author is Tin Lun Lam, email: tllam@cuhk.edu.cn and Dan Zhang, email: dan.zhang@polyu.edu.hk}
}

\begin{document}

\maketitle
\newcommand{\SLG}{\text{SLG}}
\newcommand{\AB}{\text{AB}}
\newcommand{\AC}{\text{AC}}
\newcommand{\AD}{\text{AD}}
\newcommand{\BD}{\text{BD}}
\newcommand{\BC}{\text{BC}}
\newcommand{\CD}{\text{CD}}
\newcommand{\BA}{\text{BA}}
\newcommand{\CA}{\text{CA}}
\newcommand{\DA}{\text{DA}}

\newcommand{\Outer}{\text{Outer}}
\newcommand{\Inner}{\text{Inner}}
\newcommand{\Middle}{\text{Middle}}
\newcommand{\A}{\text{A}}
\newcommand{\B}{\text{B}}
\newcommand{\C}{\text{C}}
\newcommand{\D}{\text{D}}
\newcommand{\dk}{\text{dk}}
\thispagestyle{empty}
\pagestyle{empty}

\begin{abstract}
Modular Self-Reconfigurable Robot (MSRR) systems are a class of robots capable of forming higher-level robotic systems by altering the topological relationships between modules, offering enhanced adaptability and robustness in various environments. This paper presents a novel MSRR called MODUR, featuring dual-level reconfiguration capabilities designed to integrate reconfigurable mechanisms into MSRR. Specifically, MODUR can perform high-level self-reconfiguration among modules to create different configurations, while each module is also able to change its shape to execute basic motions. The design of MODUR primarily includes a compact connector and scissor linkage groups that provide actuation, forming a parallel mechanism capable of achieving both connector motion decoupling and adjacent position migration capabilities. Furthermore, the workspace, considering the interdependent connectors, is comprehensively analyzed, laying a theoretical foundation for the design of the module's basic motion. Finally, the motion of MODUR is validated through a series of experiments.

\end{abstract}

\section{Introduction}

\begin{figure*}[thbp]
\centerline{\includegraphics[width=\linewidth]{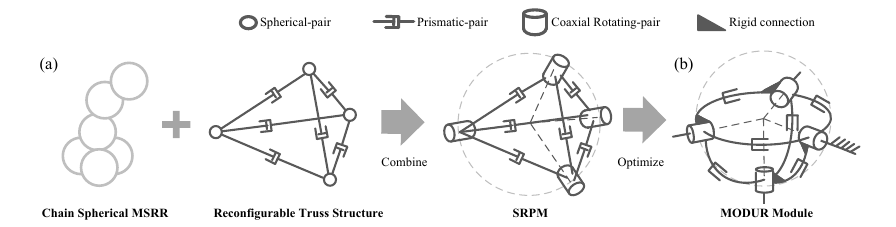}}
\caption{Schematic diagram of integrating reconfigurable mechanisms into MSRRs to derive our new conceptual model. (a) The combination of the common spherical chain MSRR and the reconfigurable truss structure, with the latter constrained within a spherical surface, forms the SRPM. (b) Simplified model, where the connectors at the vertices are rigidly connected to the associated SLGs.}
\label{fig1}
\end{figure*}

\begin{figure*}[thbp]
\centerline{\includegraphics[width=\linewidth]{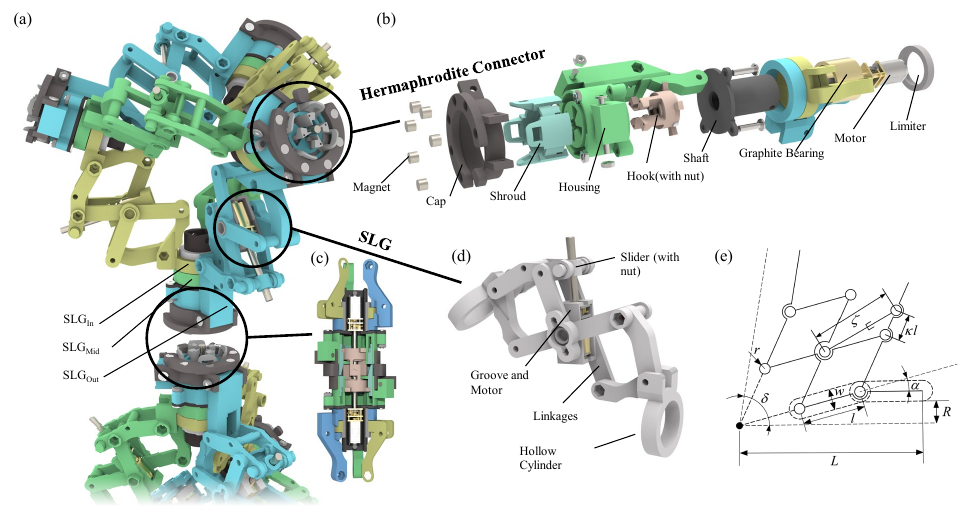}}
\caption{The specific mechanical structure design of the MODUR module. (a) The overall structure, which is composed of SLGs and connectors. (b) Exploded view of the connector. (c) Cross-sectional view when the connectors are connected. (d) The composition of the SLGs. (e) The planar design diagram of the SLG.}
\label{fig2}
\end{figure*} 

Modular self-reconfigurable robots (MSRR) are a fascinating category of robotic systems that have garnered significant attention from researchers in recent years\cite{belke2017mori,zhao2022snailbot,yim2000polybot,suh2002telecubes}. As the name suggests, MSRRs consist of two elements: modularity and self-reconfiguration. The former explains that the robot is composed of multiple identical units that can replace, combine and interact with each other\cite{romanishin2013M-blocks,castano2000conro}; the latter indicates that the robot can autonomously change the topological relationship among the units to form various configurations\cite{kurokawa2008M-TRANIII, spinos2017vtt, liang2020freebot}. For clarity, we refer to the units as \textit{modules}. Compared to rigid robots designed for specific tasks, the repetitive structures of MSRRs simplify control design and enhance robustness through interchangeability. As a result, they exhibit better adaptability to various environments and greater versatility in handling different tasks \cite{yim2007modular}.

In the theory of mechanisms and machines, there is a research field similar to self-reconfiguration, called reconfigurable mechanism (RM)\cite{PM-tian2021structure}, which can change the number or type of mobility by altering the number of effective links and the characteristics of kinematic joints\cite{tian2021design}. It can also be referred to as a metamorphic mechanism, which emphasizes the transformation of the entire mechanism across multiple modes to meet different motion requirements\cite{kang2022multiple}.

Currently, there are many advanced MSRRs, the modules of which are mostly spherical or cubic, and either occupy lattice cells or form chains, such as Roombot\cite{sproewitz2009roombots} and SMORES\cite{liu2023smores}. These modules have compact structures with connectors arranged on their surrounding faces. They rely on built-in joints to change the position of the connectors, enabling movement between modules. However, existing MSRR systems still have areas for improvement. The first is the connector motion decoupling capability, and the second is the adjacent position migration capability. To describe these two capabilities, we assume that modules form tree structures after being connected. Taking Roombot as an example, its connectors are located on the six faces of a cube and are divided into two groups, with three connectors in each group that are coupled. Therefore, for any given Roombot module, three of its child modules are coupled, while the other two are coupled with the parent module. In this case, moving a child module affects the other two coupled child modules. Additionally, the child module cannot disconnect from the parent module and reconnect with other sibling modules or the grandparent module. 

This paper introduces a novel MSRR: MODUR, a modular dual-reconfigurable robot, which is inspired by applying the concept of reconfigurable mechanisms to the modules of MSRRs\cite{pieber2018adaptive}. The modules with reconfigurable capabilities feature connectors that can move independently, overcoming the common problem of coupled connector motion present in existing MSRRs. Additionally, the modules can easily transport child modules to adjacent positions through their reconfiguration, providing MODUR with an excellent adjacent position migration capability.

MODUR is composed of four connectors and six scissor linkage groups (SLGs). The connectors feature a hermaphrodite design that combines magnetic and mechanical connections. The SLGs are six-bar structures functioning as a type of Remote Center Motion (RCM) mechanism \cite{RCM-li2024family}, serving as a P pair. Unlike MSRRs with serially arranged joints, the joints of MODUR are arranged in parallel. This arrangement allows MODUR to stand out from chain-type MSRRs, exhibiting both chain-like and lattice-like characteristics. 

This paper is organized as follows. Section II reviews related work on MSRR and RM as well as their features. Section III introduces the design concept and mechanical structure of MODUR. Section IV analyzes the workspace of the connectors under different conditions, and based on this, the basic motion is proposed. Section V presents the implementations and experiments of MODUR's prototype. The last section concludes the paper and discusses the future work.

We summarize our contributions as follows:
\begin{enumerate}
    \item A novel MSRR is proposed. This structure decouples the motion between connectors(corresponding child-modules) and provides a wide range of motion for the child-modules, facilitating more operations and 3D continuous self-reconfiguration.
    \item The workspace of the connectors is analyzed, and the basic motion of MODUR is proposed based on this, which serves as the foundation for reconfiguration.
    \item The mechanical design and basic motion capabilities have been validated through a series of experiments.
\end{enumerate}

\section{Mechanism Design of MODUR module}

This paper aims to design an MSRR module capable of adjacent position migration and decoupling between connectors, thereby enhancing the efficiency and independence of module motion. To achieve this, the design of the module incorporates the concept of reconfigurable mechanisms. The chain spherical MSRR exhibits advantages in terms of inter-module motion and spatial integration. The reconfigurable truss structure is a mechanism commonly used in parallel robots for its good stiffness and load-bearing capabilities. We integrate these two structures by constraining the vertices of the truss on a spherical surface, resulting in a spherical reconfigurable parallel mechanism (SRPM), as shown in Fig. 1(a). The SRPM has 9 degrees of freedom (DOF). To simplify control, the coaxial R-pair at each vertex is rigidly fixed to one of its associated edges. This leads to our final design structure, the MODUR module. On the one hand, MODUR exhibits the foundational design principles of existing MSRRs, forming robot systems with diverse configurations through reconfiguration. On the other hand, the MODUR module functions as a reconfigurable mechanism, capable of locking specific DOFs to achieve the desired reconfiguration capabilities.

Based on the schematic diagram in Fig. 1(b), the mechanical design of the MODUR module is shown in Fig. 2(a). It primarily comprises two key aspects: a compact and hermaphrodite connector, serving as the foundation for reconfiguration between modules, and a linkage structure named SLG, allowing the connectors at both ends to form an arc-shaped prismatic pair. SLGs are arranged along the six edges of a tetrahedron, with the corresponding connectors positioned at the four vertices. The six SLGs are not entirely identical, and they are classified into three types based on the structural differences of hollow cylinders: $\SLG_\Inner$ (yellow), $\SLG_\Outer$ (green), and $\SLG_\Middle$ (blue). These three types are arranged in a counterclockwise order at the coaxial position. The genderless connectors are fixed on the $\SLG_\Outer$s.

\subsection{Connector}
The design of connectors is crucial for the MSRR system. A qualified connector should have sufficient connection strength, precise alignment, and the ability to connect unilaterally, that is, in a hermaphrodite or genderless manner. In our design, it is also essential to consider the size of the connector. A compact design significantly enhances the workspace of the connector itself, thereby improving the overall motion performance. Similar to Higen \cite{parrott2014higen}, our designed hermaphrodite connectors achieve connection through a screw-in hook mechanism with a resolution of 90°, allowing them to be mated either in parallel or perpendicularly. However, our connectors are more compact, with a diameter of only 33 mm, and incorporate magnets with weak attractive forces for pre-alignment before connection. The specific components are shown in Fig. 2(b).

Fig. 2(c) depicts the cross-sectional view when the two connectors are connected. Before the connection, the DC gear motor, coupled with a lead screw, generates thrust to move the hook, which has a nut embedded during manufacturing, closer to its counterpart. Once the hooks come into contact and further forward movement is restricted, the motor switches to generate rotational motion \cite{nishimura2023三种抓手}, thereby locking the hooks together. In the connected state, the motors enter the third phase, continuing to provide thrust and preventing the hooks from rotating in the opposite direction through friction. Additionally, the shroud will interlock tightly to prevent the entire connector from rotating loose.

\subsection{Scissor linkage groups}
The scissor linkage groups, named SLGs, serve as the driving component of the entire module and were developed to link and mobilize connectors. In our design, after comprehensively considering the potential interference among SLGs, a radial six-bar structure, a classic RCM mechanism, is designed and modeled, as shown in Fig. 2(d). This six-bar structure is driven by a lead screw motor, with the six bars distributed across four different planes to prevent interference. To achieve the basic motion between modules, the SLG must expand within a range of at least 60° to 180°. The planar design diagram of the SLG is presented in Fig. 2(e), where the parameters are calculated using the following equations: 

\begin{equation}
{{\delta }_{\text{col}}}={{\delta }_{\min }}-2\alpha \\
\label{eq1}
\end{equation}

\begin{equation}
l=\left\lfloor \frac{r+\frac{w}{2}}{2\sin \left( \frac{{{\delta }_{\text{col}}}}{2} \right)} \right\rfloor +\lambda \\
\label{eq2}
\end{equation}

\begin{equation}
L=l+l\cos \left( \alpha  \right)+l\cos \left( 2\alpha  \right) \\
\label{eq3}
\end{equation}

\begin{equation}
R=l\sin \left( \alpha  \right)+l\sin \left( 2\alpha  \right)-\frac{w}{2} 
\label{eq4}
\end{equation}
where $\delta_\text{min}$ denotes the minimum value of $\delta$, while $\delta_\text{col}$ represents the value of $\delta$ at the moment when collision interference occurs.

Finally, the definitions and values of the selected parameters are listed in TABLE I. The hollow cylinders on both sides of the SLG are combined coaxially to embed the hermaphrodite connectors.

\begin{table}[t]
\caption{The symbols with their definitions and values.}
\begin{center}
\begin{tabular}{c p{5cm} c} 
\toprule
\textbf{Symbols} & \textbf{Definitions} & \textbf{Values} \\
\midrule
r   & The radius of the hinge holes between rods & 3mm \\
w   & The width of the rod & 9mm \\
l   & The length of the rod & 31mm \\
$\alpha$ & The deflection angle of the edge rod & 20° \\
$\lambda$ & Interference adjustment length & 7mm \\
$\kappa$ & Lever arm adjustment factor & 0.38 \\
$\zeta$ & The length of the drive rod  & \textgreater31mm \\
R   & The diameter of the connector & 16.7mm \\
L   & The radius of the overall SLG & 98.26mm \\
$\delta$ & The angle between the rotation centers of the two connectors connected by the SLG & $\geq$ 60°\\
$l_\dk$ & The distance between the two connected connectors & 8mm \\
\bottomrule
\end{tabular}
\label{tab1}
\end{center}
\end{table}

\section{Motion Analysis}
\subsection{Kinematic Model}
The Kinematic model of MODUR and coordinate system are established as shown in Fig. 3. The four connectors are represented by four points A, B, C, and D on a sphere. The connector at point D is connected to the parent module and is referred to as the input connector, with point D also being called the input point. The point sharing $\SLG_\Outer$, $\SLG_\Middle$, and $\SLG_\Inner$ with D is A, B, and C, respectively, called output point. During each motion process, there can only be one input point, and the remaining points are output points. In the abstract model, three output points can be distinguished by \eqref{eq5}, where E is the centroid of the triangle $\Delta$ABC.

\begin{equation}
\boldsymbol{CE} \cdot (\boldsymbol{AB} \times \boldsymbol{AD}) > 0\label{eq5}
\end{equation}

At point D, $\boldsymbol{z_D}$ is defined along the direction pointing from the parent module to D, while $\boldsymbol{z_A}$, $\boldsymbol{z_B}$, and $\boldsymbol{z_C}$ are defined along the directions pointing to their respective virtual child modules. The plane perpendicular to $\boldsymbol{z_D}$ and passing through the center of the sphere O is designated as the horizontal plane \( \Pi_h \). $\boldsymbol{x_D}$ and $\boldsymbol{x_A}$ are defined on the plane containing \( \text{SLG}_{AD} \), while $\boldsymbol{x_B}$ and $\boldsymbol{x_C}$ are defined on the plane containing \( \text{SLG}_{BC} \). The coordinate system of the sphere center \{O\} can be obtained by translating the coordinate system \{D\}. The positions of points A, B, and C are established in \{O\} and are determined by the parameters \( \varphi_A \), \( \varphi_B \), \( \varphi_C \), \( \theta_B \), and \( \theta_C \). Here, \( \varphi \) represents the angle relative to the horizontal plane \( \Pi_h \), with a positive value indicating a direction away from D, and \( \theta \) represents the angle relative to $\boldsymbol{x_O^+}$, with a positive value indicating a counterclockwise rotation around $\boldsymbol{z_O}$. The coordinate system \( \{D^+\} \) is established at the input point of the child module and includes a radial deflection angle \( \phi \), which is determined during the connection process and cannot be altered. This angle \( \phi \) is a multiple of 90° and influences the orientation of the child-module coordinate system.
\begin{figure}[t]
    \centering
    \includegraphics[width=0.8\linewidth]{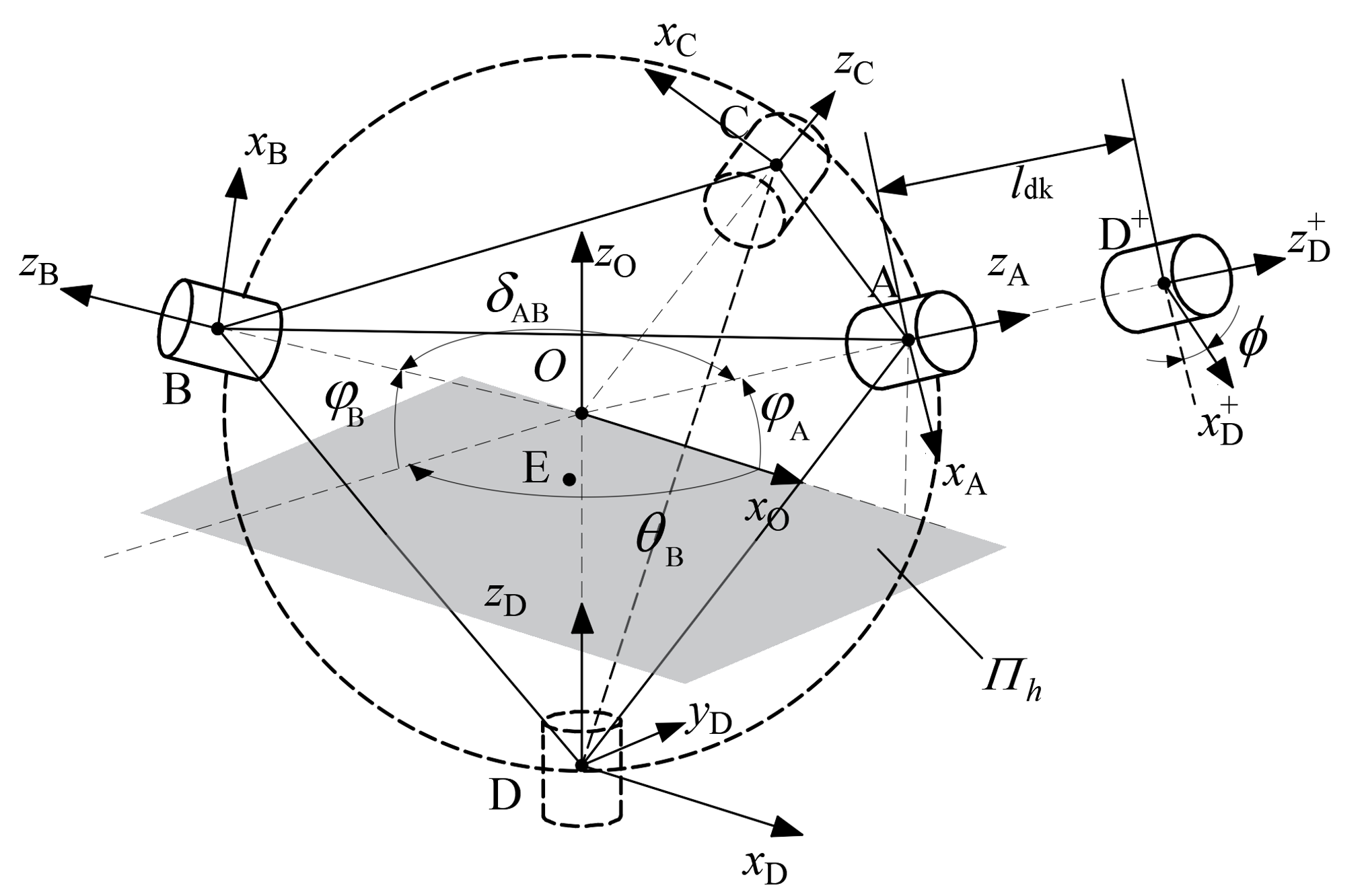}
    \caption{The abstract model of MODUR, where coordinate frames are established, and parameters are labeled. }
    \label{fig3}
\end{figure}

\begin{equation}
\cos \phi_C =
\begin{cases}
\displaystyle
\frac{\cos \varphi_{B} \cos \left( \theta_{C} - \theta_{B} \right) - \cos \delta_{BC} \cos \varphi_{C}}{\sin \delta_{BC} \sin \varphi_{C}}, \\ 
\quad\quad\quad\quad\quad\quad\quad\quad\quad\quad\quad\quad \text{if } \varphi_C \neq 0  \\[10pt]
\displaystyle
\frac{\cos \varphi_B - \cos \delta_{BC} \cos (\theta_C - \theta_B)}{\sin \delta_{BC} \sin (\theta_C - \theta_B)}, \\ 
\quad\quad\quad\quad\quad\quad\quad\quad\quad\quad\quad\quad \text{if } \varphi_C = 0 
\end{cases}
\label{eq6}
\end{equation}

In our latest study \cite{gu2024moparas}, we performed a comprehensive kinematic analysis of a 3-RPR spherical mechanism that represents MOPARAS's structural configuration when three connectors are fixed with their corresponding DOFs constrained. In this new model, similarly, we can solve for the position parameters $\varphi$ and $\theta$ at three output points using the same numerical methods. Differently, since our connectors are fixed to the SLG, the orientation parameter $\phi$ (e.g., $\phi_C$) can be determined through (6).

\subsection{Workspace analysis}

\begin{figure*}[t]
\centerline{\includegraphics[width=\linewidth]{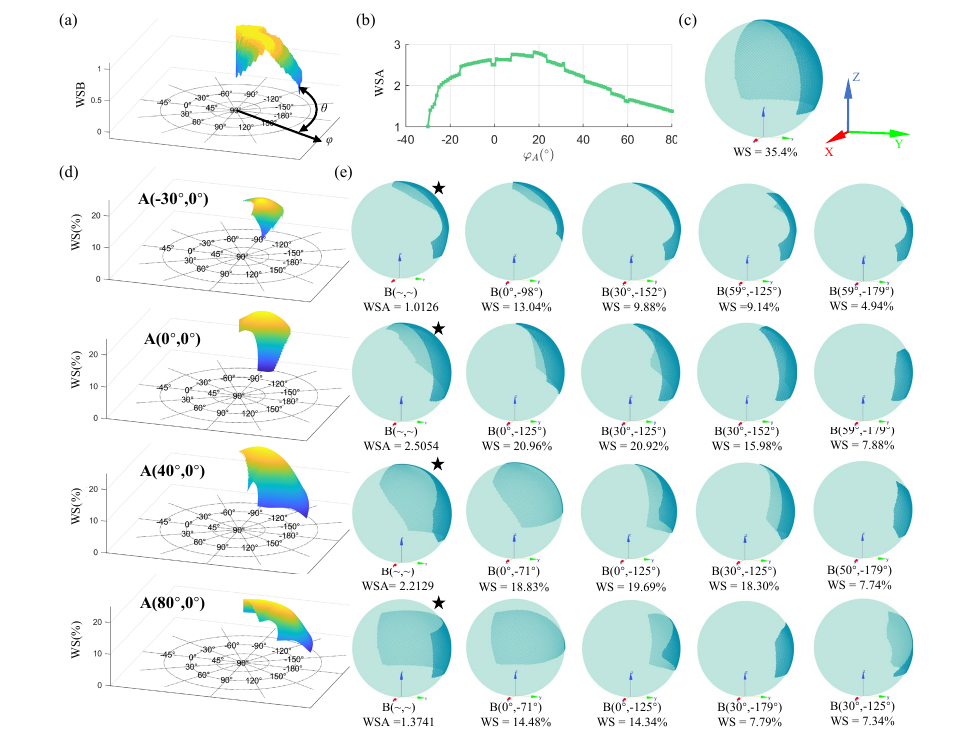}}
\caption{The workspace under different conditions where some connectors are in a connected state. (a) WSB reflects the impact of the positions of connector B on the workspace of connector C. (b) WSA reflects the impact of the positions of connector A on the workspace of connector C. (c) Total workspace diagram with no constraints on A and B. (d) The heatmap showing the impact of different B positions on the workspace of C for selected values of A. (e) The workspace diagram of connector C for all positions of connector B (with star) and selected positions of B with A at a specific position (without star), corresponding to (d).
}
\label{fig5}
\end{figure*}

Based on the aforementioned kinematics, we analyze the workspace of the module's three output connectors. This analysis is essential because the connectors' motions are kinematically coupled — the workspace of each connector varies with the positions of the others. The workspace is mainly affected by the following three constraints:
\begin{itemize}
    \item $\delta$ needs to satisfy being between 60° and 180°.
    \item The SLGs should not interfere at the connectors.
    \item Opposite SLGs (e.g., $\SLG_{\AC}$ and $\SLG_{\BD}$ are Opposite) should not intersect.
\end{itemize}

For clarity, connectors A, B, C, and D are hereafter abbreviated as A, B, C, and D, respectively. For A, since A and D are both fixed on the $\SLG_\AD$ and according to the geometric condition of constraint 1, we can determine that its workspace is a $120^\circ$ arc.  

Then, C is chosen as the subject of analysis, as the cases for B and C are similar. During the analysis, we assume that A and B are not in a connected state, so A and B can take any position within the constraints. Constraints 1 and 2 are guaranteed by structural design. When analyzing C, the main limitation comes from constraint 3. The Möller–Trumbore ray tracing method is used to exclude cases where opposite edges intersect \cite{moller2005fast}.
 
Considering that the position of A is influenced by only one variable, we simplify the analysis by fixing the position of A at specific values and then examining how the workspace of C changes as the position of B varies. The workspace of C is quantified by the percentage of the sphere’s surface area it occupies, denoted as WS. By enumerating multiple positions of B, the sum of all corresponding WS values is referred to as WSA. This sum, WSA, serves as a measure of the total workspace available to C as the position of A changes across different fixed values. Additionally, we define WSB as the proportion of A’s positions where the workspace of C is non-zero for a given B position. For example, if we fix 10 positions of A, and in 6 of them, B’s position results in a non-zero workspace, then WSB would be 0.6. WSA and WSB reflect the impact of the positions of connectors A and B on the workspace of connector C, with the results shown in Fig. 4(a) and Fig. 4(b), respectively.

Fig. 4(c) shows the total workspace of C for all possible positions of A and B. This indicates that if A and B are not in a connected state, meaning their positions can be adjusted to accommodate the movement of C, the workspace of C can approach half of the sphere's surface, with WS$=35.4\%$. The heatmap in Fig. 4(d) exemplifies the changes in the workspace size of C as B moves for different positions of A, and Fig. 4(e) correspondingly plots the workspace diagrams of C for different configurations. In each row of these diagrams, the first diagram, marked with a star, represents the total workspace of C when A remains fixed while B moves freely. The second to fifth diagrams illustrate the progressive variations in C’s workspace as the positions of A and B change.

\subsection{Basic Motion}\label{AA}

\begin{figure}[t]
\centerline{\includegraphics[width=\linewidth]{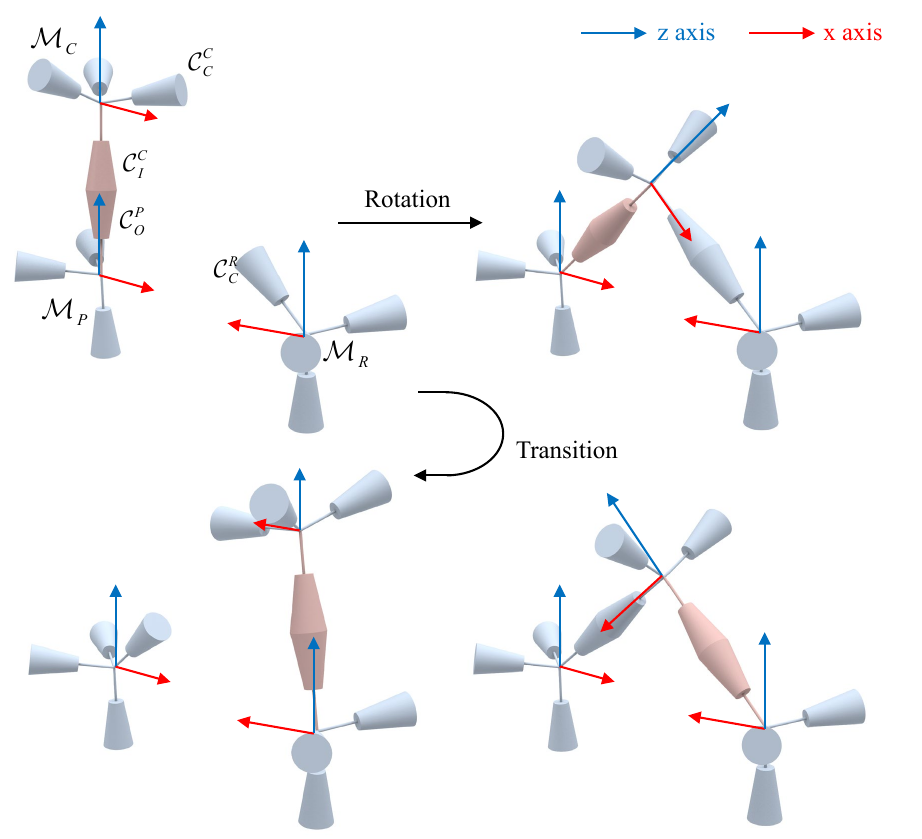}}
\caption{Basic motion between modules: Rotation motion and transition motion.}
\label{fig6}
\end{figure}


The workspace analyzed above forms the foundation for the motion analysis of the connector and its corresponding child module. Within the workspace, the child module can perform some basic motions. Thanks to the folding range of the SLG mechanism, MODUR has two basic motions, shown as Fig. 5: one is rotation and the other is transition. Rotation does not require connection or disconnection between modules, whereas transition does. For the sake of clarity, we define the module waiting for transition as the current module, named $\mathcal{M}_C$, whose input connector (defined in Section III) is denoted as $\mathcal{C}^{C}_{I}$, and its participating connector is denoted as $\mathcal{C}^{C}_{C}$. The parent module is denoted as $\mathcal{M}_P$, and its output connector connected to $\mathcal{C}^{C}_{I}$ is denoted as $\mathcal{C}^{P}_{O}$. The module receiving it is denoted as $\mathcal{M}_R$, whose participating connector is denoted as $\mathcal{C}^{R}_{C}$.

To realize these two basic motions, two conditions need to be satisfied during the motions:

\textbf{Condition 1:} The condition for a module to perform rotation motion is that the connector of the parent module must have available workspace in the direction of rotation.

\textbf{Condition 2:} The conditions for transition motion are as follows: 

\begin{itemize} 
\item $\mathcal{M}_C$: Along the $\boldsymbol{u}$ direction, there are free connectors with $\varphi=-30^\circ$. 
\item $\mathcal{M}_P$: $\mathcal{C}^{P}_{O}$ has a workspace at the position forming a 120° angle with $\boldsymbol{z}^+$.
\item $\mathcal{M}_R$: $\mathcal{C}^{R}_{C}$ has a workspace at the position forming a 30° angle with $\boldsymbol{z}^+$.

\item $\mathcal{C}^{R}_{C}$ and $\mathcal{C}^{C}_{C}$ are on the same plane.

\end{itemize}
After the transition, the coordinate system {O} is re-established based on the new connection configuration. For example, when point A becomes the input point connected to a new module, the $\boldsymbol{x}$ and $\boldsymbol{z}$ axes will be re-established, with A as the new origin (D).


\section{Implementations and experiments}
Based on the aforementioned design and theoretical analysis, this section describes the hardware implementation and motion experiments of our functional prototype, MODUR, shown in Fig. 6(a). The entire structure is fabricated using 3D printing with PLA material. Due to the decentralized nature of the overall structure, designing integrated electronics is challenging and is not included in the scope of this study. MODUR is controlled by an Arduino Mega2560, which interacts with the host computer. To provide sufficient motor drive, a 12V power supply is utilized, and two six-channel 5V to 12V relays are employed to receive control signals from the Arduino, enabling the forward and reverse operation of the motors. The TCA9548A is used to resolve I2C slave address conflicts, allowing it to receive angle signals from six angle sensors. MODUR is equipped with 12 N20 geared motors rated at 12V, six of which drive the connection and disconnection of connectors, while the other six drive the contraction and extension of the SLG. Furthermore, six AS5600 sensors are positioned within the SLGs, as illustrated in Fig. 6(b), to measure the deployment angles of the SLGs. However, due to insufficient stiffness of the machining materials, there is always an unfold angle error $\delta_e$  (within 9°) in the SLGs. Additionally, the structural design shown in Fig. 2(e) introduces a deviation $\alpha$ between the sensor-measured angle $\delta_s$ and the actual angle $\delta$. The actual angle $\delta$ can be derived using Equation (7).
\begin{equation}
\delta = \delta_s+2\alpha+\delta_e
\label{eq8}
\end{equation}

\begin{figure}[t]
\centerline{\includegraphics[width=\linewidth]{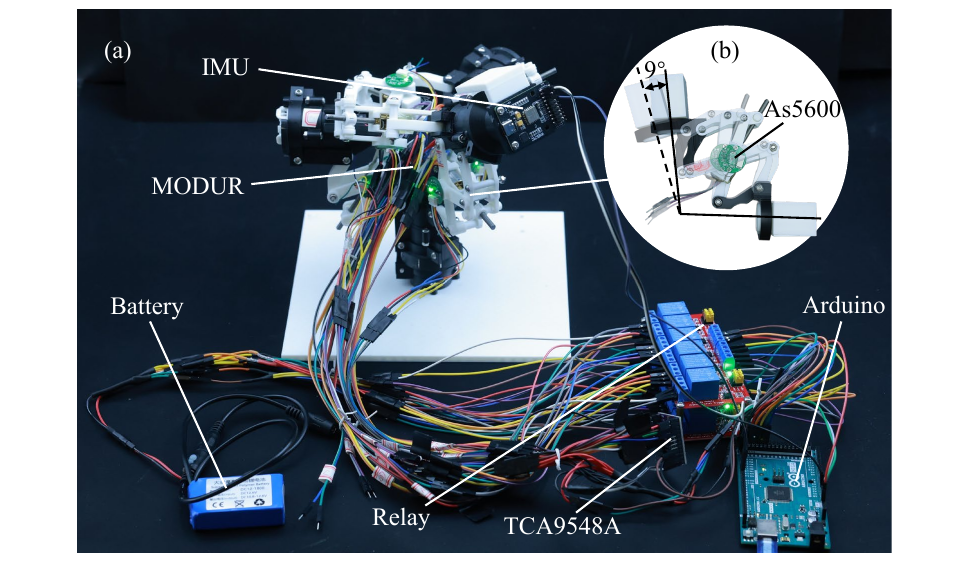}}
\caption{The hardware implementation. (a) The prototype, MODUR, is equipped with basic components such as drives and power. (b) An unfold angle error $\delta_e$ (within 9°) in the SLGs.}
\label{fig7}
\end{figure}

\subsection{Manipulation}
Rotation motion is one of the basic motions and serves as the foundation for a module to transport or move its child modules. When modules are connected in series, the current module can manipulate all its child modules. In this case, the rotation motion can be equivalent to the revolute joint of a serial robot.

\subsubsection{Control strategy}
\begin{figure}[t]
\centerline{\includegraphics[width=\linewidth]{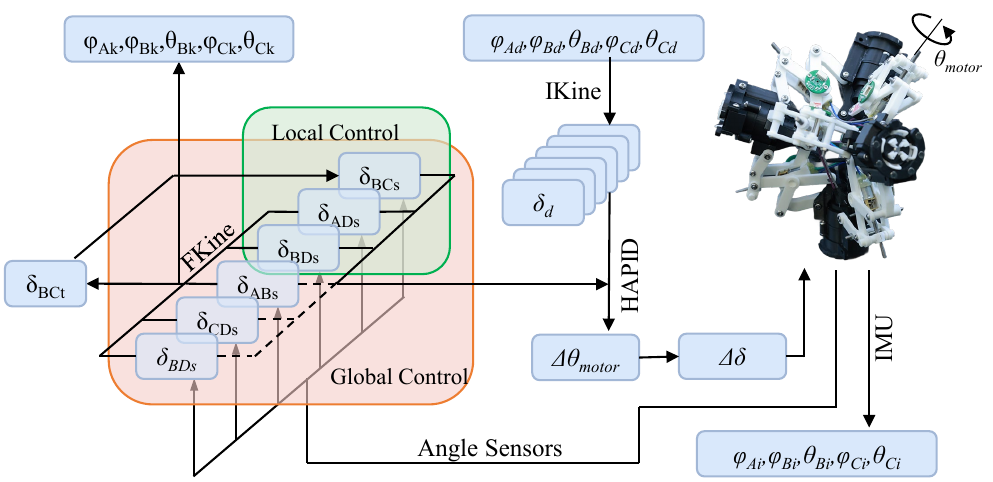}}
\caption{Flowchart of the control strategy}
\label{fig8}
\end{figure}


\begin{figure}[t]
\centerline{\includegraphics[width=\linewidth]{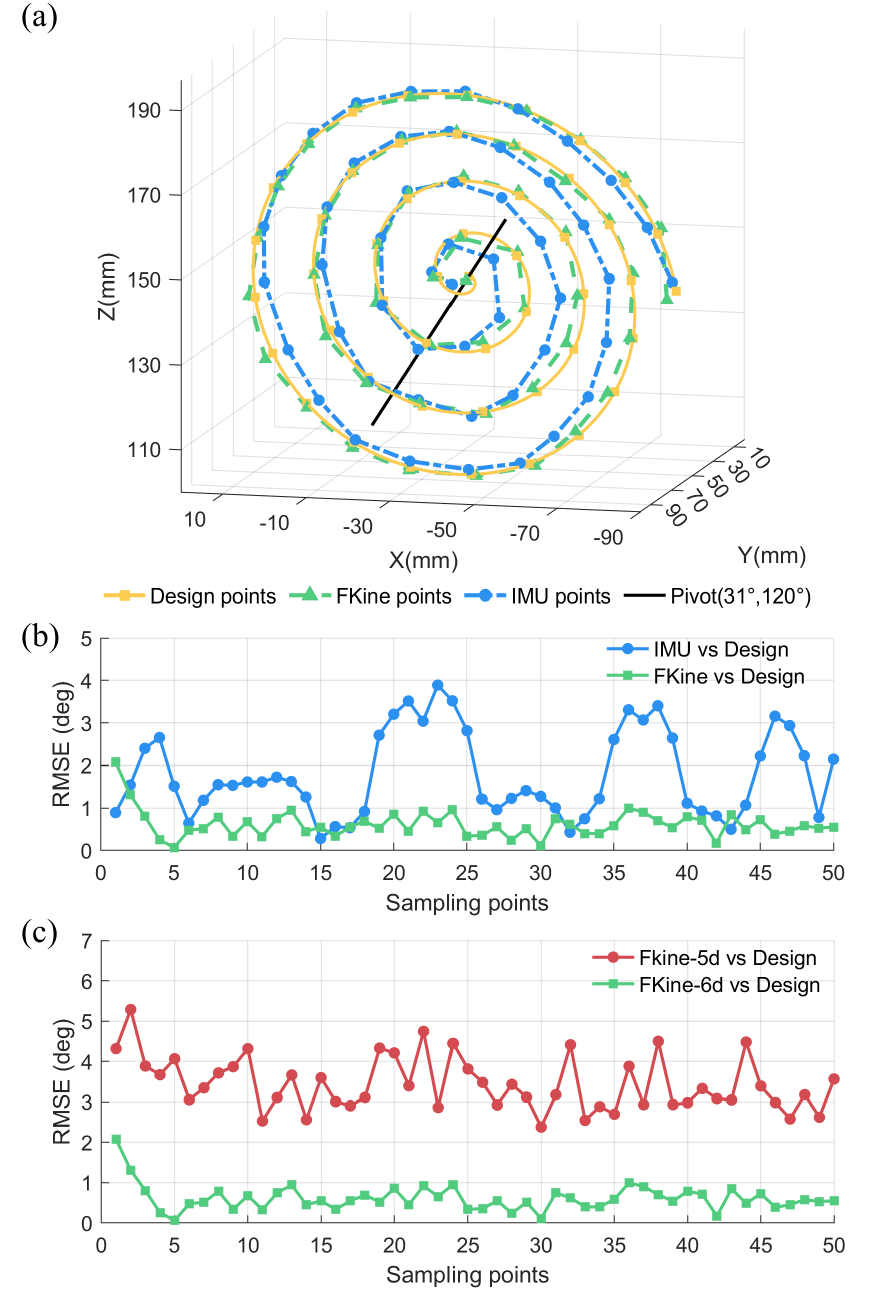}}
\caption{Experimental results of trajectory tracking for a module connector.  
(a) Designed trajectory, SLG angle sensor-derived trajectory, and IMU-measured trajectory. 
(b) Comparison of position deviation between the trajectory derived from the angle sensor measurements and the IMU-measured trajectory.  
(c) Comparison of position deviation with and without redundant actuation.}
\label{fig9}
\end{figure}

The entire MODUR is equipped with 6 motors for actuation, despite having only five DOF, which offers two advantages.:

\begin{itemize}
\item The introduction of redundant drives can effectively prevent the structure from reaching singular positions.
\item Constraining the sixth SLG can effectively reduce the impact of the unfold angle error $\delta_e$.
\end{itemize}

However, the redundancy in actuation may lead to redundant force conflicts. To address this, a hierarchical approximation PID control method (HAPID) is proposed. In the high-level control layer, the convergence goal is decomposed into multiple-stage goals, ensuring that no SLG’s unfold angle change exceeds the unfold angle error limit during each PID execution. Excessive changes in the unfold angle may create internal force conflicts, leading to insufficient motion and causing the structure to converge to a suboptimal configuration. An intuitive decomposition method is adopted in this paper:

\begin{equation}
steps=\lfloor \max\{ \frac{\Delta\delta_i}{\delta_e} \}\rfloor,i=AB,...,CD 
\label{eq9}
\end{equation}

\begin{equation}
\bm\delta_{sgi} =  \bm\delta_{d} + \frac{k\Delta \bm\delta}{steps}  ,k = 1,...,steps
\label{eq10}
\end{equation}

In the low-level control layer, PID control is applied based on stage targets. To improve convergence speed, a loose convergence threshold PID control can be used for all steps except the final one.

Fig. 7 illustrates the control flowchart based on the aforementioned strategy. In the diagram, Fkine and Ikine denote the forward and inverse kinematics functions, respectively. As described in Section IV-A, when three connectors remain stationary while one is actively controlled, the entire mechanism can be simplified to a 3-RPR parallel structure. Under conditions where high overall control accuracy is not required, this single-connector control strategy—referred to as local control—can be employed. Here, three motors drive a 2-DOF mechanism, achieving faster execution compared to global control while trading off precision for speed.

\subsubsection{Accuracy}

The experimental platform is shown in Fig. 6. To verify the advantages of the aforementioned redundant drive control method and to examine the differences between adding sensors at the SLGs and end connectors, a helical trajectory was designed for the connector with an Inertial Measurement Unit (IMU), as shown in Fig. 8(a). The trajectory passes through point (40°, 160°) with (30°, 120°) as the center of rotation, and 50 evenly spaced points were selected as drive goals, marked in yellow in the figure. The green curve and marks represent the trajectory and control points obtained using global control, while the blue curve and marks represent the trajectory and control points calculated based on the data from IMU. The root mean square error (RMSE) of the connector position was used to measure the deviation distance between the global control trajectory and the designed trajectory, as well as between the IMU-measured trajectory and the designed trajectory, as shown in Fig. 8(b). The first RMSE indicates that the control method generally ensures the position deviation remains within 1°. A comparison of both RMSEs demonstrates that the control effects of adding sensors at the SLGs and end connectors are similar. However, a significant deviation in the horizontal position is observed, likely due to the pulling force exerted by the Dupont wires.

A comparative experiment was conducted by removing the motor in $\SLG_\BC$ to compare the control effects of redundant and non-redundant drives. The experimental results, shown in Fig. 8(c), indicate that the non-redundant drive yields poorer control performance. This may be due to the tendency of $\SLG_\BC$ to lean towards the side affected by gravity, resulting in an inaccurate end position.

\subsection{3D Reconfiguration}

\begin{figure*}[thbp]
\centerline{\includegraphics[width=\linewidth]{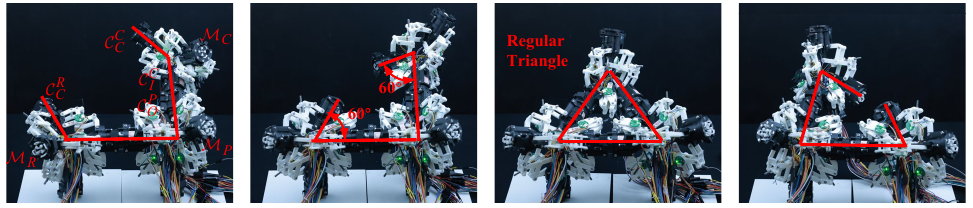}}
\caption{Demonstration of MOUDR's transition with the connection already established between $\mathcal{M}_R$ and $\mathcal{M}_P$. $\mathcal{M}_R$ and $\mathcal{M}_P$ are both connected to the base and each other, meaning each has a connector with a fixed position.  $\mathcal{M}_C$ transitions between them. (a) Initial state; (b)  $\mathcal{M}_C$ and $\mathcal{M}_R$ adjust the connector awaiting connection to the designated position; (c) Establishing a new connection; (d) $\mathcal{M}_C$ disconnects from $\mathcal{M}_P$.}
\label{fig13}
\end{figure*}

No additional demonstration is required for the rotational motion, as it is involved in validating the transitional motion. An experiment was conducted here to validate the transition capability of the modules with the connection already established between $\mathcal{M}_R$ and $\mathcal{M}_P$, to demonstrate the ability to transition between modules.

The experiment is shown in Fig. 9. In the initial state, $\mathcal{M}_C$ can be in any position and orientation. $\mathcal{M}_R$ is placed at a distance of 2L from $\mathcal{M}_P$ (L is given in Table I). When the connection begins, $\mathcal{C}^{C}_{C}$ will rotate to form a 60° angle with $\mathcal{C}^{C}_{I}$. $\mathcal{C}^{R}_{C}$ will then rotate to form a 60° angle with the line between $\mathcal{M}_P$ and $\mathcal{M}_R$. After that, $\mathcal{M}_P$ will manipulate $\mathcal{M}_C$ to connect to $\mathcal{M}_R$. The centers of the three modules will form an equilateral triangle, indicating that the three modules are connected to each other. Finally, $\mathcal{M}_C$ will disconnect from $\mathcal{M}_P$.

When multiple MODURs are chained together to form a robotic arm, the end MODUR can be transported to a further position and connected to another MODUR, as long as condition 2 in Section IV-C is satisfied. 


\section{conclusion}
This paper proposes a novel MSRR called MODUR, which integrates reconfigurable mechanisms into the modules of MSRRs. This endows the MODUR modules with enhanced adjacent position migration capability and connector motion decoupling capability. As a result, the child modules of the MODUR module do not interfere with each other and can move to adjacent positions topologically. This serves as the cornerstone for achieving 3D continuous reconfiguration. The workspace is analyzed based on the constraints of each connector, providing guidance for the module's motion. Next, we conducted experiments to validate the ability to manipulate the modules. The results show that the redundant actuation method based on HAPID offers higher precision compared to the non-redundant approach, and the absolute position measured by the IMU is also generally well-controlled during implementation. Finally, we demonstrated the 3D reconfiguration capability of MODUR through physical experiments.

Our future work will focus on three main aspects. The electronic design of the robot will be improved to make the overall structure more compact and capable of performing basic actions. Secondly, we aim to explore the advantages of multiple modules in tasks such as grasping and manipulation. Additionally, a motion planning framework based on 3-D reconfiguration capability needs to be developed.

\addtolength{\textheight}{-12cm}   


\bibliographystyle{IEEEtran}
\bibliography{references}

\end{document}